\documentclass[letterpaper,twocolumn,10pt]{article}
\usepackage{zhanggroup}

\usepackage{times}
\usepackage{latexsym}
\usepackage{microtype}
\usepackage{inconsolata}
\usepackage{graphicx}
\usepackage{amsmath}
\usepackage{amssymb}
\usepackage{amsthm}
\usepackage{xspace}
\usepackage{multirow}
\usepackage{pifont}
\usepackage{wasysym}
\usepackage[normalem]{ulem}
\newcommand{\mypara}[1]{\noindent{\bf {#1}.}}
\newcommand{\method}{LatentLens\xspace}
\newcommand{\significant}{$^*$\xspace}
\newcommand{\cmark}{\ding{51}}
\newcommand{\xmark}{\ding{55}}
\begin{document}
%-------------------------------------------------------------------------------

\date{}

\title{\bf \method: Rethinking Latent Interface in Multi-Agent Systems}

\author{
Ziqing Yang\textsuperscript{1}\ \ \
Kewei Cheng\textsuperscript{2}\ \ \
Mingjie Li\textsuperscript{1}\ \ \
Yang Zhang\textsuperscript{1}\thanks{Corresponding author.}
\\
\\
\textsuperscript{1}\textit{CISPA Helmholtz Center for Information Security} \ \ \
\textsuperscript{2}\textit{Microsoft}
}

\maketitle

%-------------------------------------------------------------------------------
\begin{abstract}
Latent interfaces let agents in a multi-agent system communicate through continuous representations rather than decoded text.
Prior work reports performance and efficiency gains, but neither separates latent communication from the agent harness nor examines its internal components, leaving its actual contribution unclear.
We propose \method, a framework that decouples the latent interface at two levels: \textit{externally} from the agent harness, to measure its contribution to overall performance (RQ1), and \textit{internally} into realignment compatibility, information accessibility, and propagation dynamics, to identify the benefits and limitations of each (RQ2).
Across seven benchmarks with Qwen3 models, adding the latent interface to the same harness does not improve and can even substantially degrade accuracy.
Internally, latent representations retain some domain information, but incompatible realignment can substantially degrade performance, and repeated latent steps can amplify representation-level failures.
Thus, while latent representations may preserve information, they do not necessarily ensure effective communication, which also requires compatible realignment, information accessibility, and stable propagation.
\end{abstract}
%-------------------------------------------------------------------------------

%-------------------------------------------------------------------------------
\section{Introduction}
%-------------------------------------------------------------------------------

Recently, large language model (LLM)-based multi-agent systems have attracted substantial interest as a way to solve complex tasks through collaboration~\cite{WBZWLZJZZLAWBW24,HZCZCWZWYLZRXWS24}.
While agents commonly exchange textual messages, recent work explores the \emph{latent interface} that bypasses intermediate token decoding and communicates through continuous representations~\cite{PLYCLYPWY24,ZYQLTLSTCHZWY26,FMZYDOW26}.
These works assume that hidden information preserved in such continuous representations can enhance reasoning and report improvements in task performance and efficiency~\cite{HSSLHWT24,ZYQLTLSTCHZWY26,DWBCZCZCY26}.

\begin{figure*}[t!]
\centering
\includegraphics[width=\textwidth]{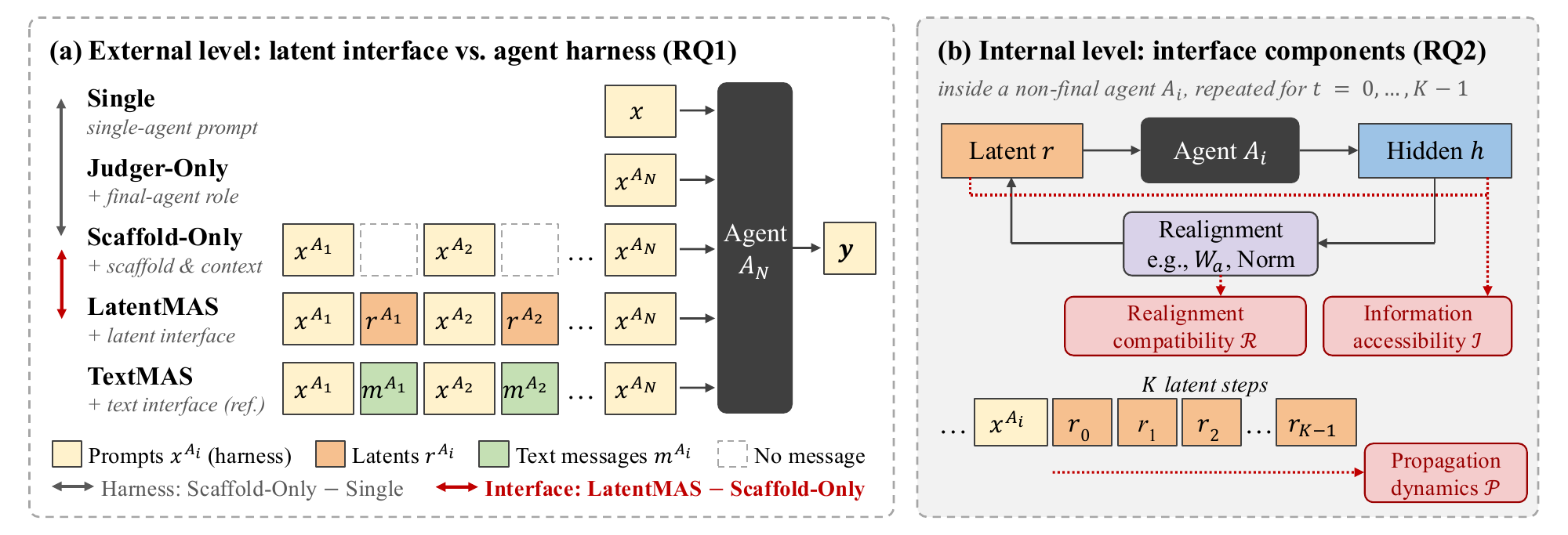}
\caption{
Overview of \method.
(a) External level: each row shows what the final agent $A_N$ receives before generating the answer $y$.
From Single to LatentMAS, each condition adds one factor, separating the harness effect from the interface effect; TextMAS utilizes the traditional text interface and serves as a reference.
(b) Internal level: in each latent step of a non-final agent, the hidden state is realigned and normalized into a latent input and fed back to the model.
$\mathcal R$ and $\mathcal I$ characterize a single step, and $\mathcal P$ characterizes how their effects evolve over $K$ steps.
}
\label{figure:latent_interface_overview}
\end{figure*}

However, improvements of a complete system do not by themselves establish the contribution of its latent interface.
On the one hand, as shown in Figure~\ref{figure:latent_interface_overview}, a multi-agent design also introduces role-specific prompts, upstream context, and additional computation, which we refer to as the \emph{agent harness}.
Consequently, outperforming a single-agent baseline does not determine whether latent inputs provide the benefit or whether the harness accounts for it.
On the other hand, the latent interface itself is not a single operation: a source hidden state is transformed into a latent input, processed by the model, and incorporated into the state and context used by later steps.
A system-level outcome, whether positive or negative, therefore does not reveal which of these components helps or hurts.
This leaves two complementary research questions:
\begin{itemize}
\item \textbf{RQ1:} \emph{What does the latent interface contribute to overall multi-agent performance when decoupled from the agent harness?}
\item \textbf{RQ2:} \emph{What are the benefits and limitations of the key components within latent interfaces?}
\end{itemize}

To answer both questions, we propose \method, a framework that decouples the latent interface at two levels.
At the \emph{external} level, \method adds harness factors cumulatively, so that the interface's contribution is measured under a matched harness (RQ1).
Its key control is Scaffold-Only, a condition we introduce that retains the upstream agents' prompts but omits latent inputs.
At the \emph{internal} level, \method follows the computation of a latent step and decomposes the interface into three connected components (RQ2):
(a) \emph{realignment compatibility} ($\mathcal R$), whether transformed hidden states are suitable as model inputs;
(b) \emph{information accessibility} ($\mathcal I$), whether the representation provides information that the receiver can effectively use;
and (c) \emph{propagation dynamics} ($\mathcal P$), how subsequent latent steps preserve, amplify, or suppress these effects.

We take LatentMAS~\cite{ZYQLTLSTCHZWY26}, a concrete instantiation of the common latent-interface pipeline, as a case study, and evaluate seven benchmarks with Qwen3-4B and 8B~\cite{T25} under default settings.
For RQ1, Scaffold-Only already outperforms the single-agent baseline, demonstrating the benefits of the harness itself.
However, LatentMAS, which introduces a latent interface into the same harness, reduces mean accuracy across seven benchmarks by 6.24\% and 17.45\% for Qwen3-4B and Qwen3-8B, respectively.
This challenges the assumption that utilizing continuous hidden states necessarily benefits reasoning.

For RQ2, removing the realignment matrix largely restores the performance of Qwen3-8B, an untied model~\cite{T25} whose realignment does not closely recover the input embeddings, suggesting that poor realignment compatibility can impair LatentMAS's performance.
Domain probes show that latent representations are close to text at the first state, but the text--latent gap widens as latent states accumulate, with finer-grained subdomain information fading faster, suggesting that latent representations carry little identifiable task-specific information.
Qwen3-4B, whose realignment is near-identity, remains below Scaffold-Only even without the matrix and degrades as latent steps increase.
Repeated latent steps can thus propagate representation-level failures.

Taken together, our findings highlight that \emph{preserving information in latent representations does not guarantee effective communication}.
A hidden state may retain task-relevant information without contributing to the receiver's reasoning: effective communication also requires compatible realignment, accessible information, and stable propagation across latent steps.
These findings motivate future work, both trained and training-free, on realignment and on preserving useful information.

Overall, our contributions are as follows.
\begin{itemize}
\item We propose \method, a two-level framework that decouples the latent interface externally from the agent harness and internally further decomposes it into three components for systematic analysis.
\item We show that the latent interface does not improve multi-agent performance once the harness is controlled.
\item We show that preserving information alone is not sufficient for latent communication: latent representations retain some domain information, yet incompatible realignment can substantially degrade performance and repeated latent steps can amplify representation-level failures.
\end{itemize}

%-------------------------------------------------------------------------------
\section{Preliminary}
%-------------------------------------------------------------------------------

%-------------------------------------------------------------------------------
\subsection{Latent Interface}
\label{section:latent_interface}
%-------------------------------------------------------------------------------

Let $f_\theta$ be a frozen LLM with vocabulary size $V$ and hidden dimension $d$.
We denote its input embedding matrix and output head by $W_{\mathrm{in}}, W_{\mathrm{out}} \in \mathbb{R}^{V \times d}$, respectively, and use row-vector notation throughout.
Given tokenized input text $x=(x_1,\ldots,x_n)$, the model processes the embeddings $E(x)=[W_{\mathrm{in}}[x_1];\ldots;W_{\mathrm{in}}[x_n]]$.
Let $h_0 \in \mathbb{R}^{1 \times d}$ denote the final hidden state at the last input position, and let $C_0$ denote the resulting KV cache.
In standard text generation, the model selects a token from the logits $h_0 W_{\mathrm{out}}^\top$ and uses its input embedding through $W_{\mathrm{in}}$ for the next step.

A \emph{latent interface} instead realigns a hidden state directly into a continuous input, bypassing discrete token selection and input token embedding.
We denote this as $r_t = \mathrm{Realign}(h_t)$, where $h_t$ is the source hidden state and $r_t$ is the realigned vector supplied as the next input embedding.
In this study, we consider a widely used realignment function~\cite{ZYQLTLSTCHZWY26,YZPQLDJTZBHZ26,YLJYKW26,LWLLQLLHYZ26}:
\begin{equation}
\label{equation:realignment_function}
r_t = h_t W_a, \text{where}\  W_a \approx  W^{\dagger}_{\mathrm{out}}W_{\mathrm{in}}.
\end{equation}
$W^{\dagger}_{\mathrm{out}}$ is a pseudo-inverse of $W_{\mathrm{out}}$, where the details can be found in Appendix~\ref{appendix:realignment}.

We define a \emph{latent step} as constructing $r_t$ and
processing it through the model:
\begin{equation}
\label{equation:latent_step}
(h_{t+1},C_{t+1})=f_\theta(r_t;C_t),
\end{equation}
where $f_\theta(r_t;C_t)$ denotes a forward pass with $r_t$ supplied through the input-embedding interface and $C_t$ as the preceding cache.
Each step adds one input position and updates the cache.
Executing $K$ latent steps injects
$R=(r_0,\ldots,r_{K-1})$ and produces the final state $h_K$ and cache $C_K$.

%-------------------------------------------------------------------------------
\subsection{Latent Interface in Multi-Agent Systems}
%-------------------------------------------------------------------------------

The latent interface could be naturally adopted in a multi-agent system, as shown in Figure~\ref{figure:latent_interface_overview}.
Formally, we use \emph{sender} and \emph{receiver} to denote the computational roles of each agent model~\cite{ZE26}.
A sender produces text/latent inputs to a receiver; then the receiver continues generation conditioned on this input context.
Each agent takes all prompts and latents generated by previous agents with its role-specific prompt as input.
Non-final agents perform $K$ latent steps while the final agent generates the final textual answer directly.
Note that $K$ is the number of latent steps per non-final agent, not the total across the pipeline.

%-------------------------------------------------------------------------------
\section{\method: Two-Level Decoupling}
\label{section:method}
%-------------------------------------------------------------------------------

\method analyzes a latent interface at two levels (Figure~\ref{figure:latent_interface_overview}).
The external level asks what the interface adds to a multi-agent system beyond its agent harness (RQ1), whereas the internal level asks how the components of the interface help or hurt (RQ2).

%-------------------------------------------------------------------------------
\subsection{External Level: Interface vs.\ Harness}
\label{section:external}
%-------------------------------------------------------------------------------

\begin{table}[t]
\centering
\caption{
Evaluation conditions at the external level.
\textbf{Role}: role-specific prompt.
\textbf{Scaffold}: upstream agents in the pipeline.
\textbf{Extra compute}: computation whose result reaches the final agent, beyond final-answer generation.
Above the line, each condition adds one factor to the one before it.
}
\label{table:effect_condition_comparison}
\setlength{\tabcolsep}{3.3pt}
\begin{tabular}{lccccc}
\toprule
& \multicolumn{2}{c}{\textbf{Harness}} & \multicolumn{2}{c}{\textbf{Interface}} & \textbf{Extra} \\
\cmidrule(lr){2-3}\cmidrule(lr){4-5}
\textbf{Condition} & \textbf{Role} & \textbf{Scaffold} & \textbf{Latent} & \textbf{Text} & \textbf{compute} \\
\midrule
Single        & \xmark & \xmark & \xmark & \xmark & -- \\
Judger-Only   & \cmark & \xmark & \xmark & \xmark & -- \\
Scaffold-Only & \cmark & \cmark & \xmark & \xmark & Prefill \\
LatentMAS     & \cmark & \cmark & \cmark & \xmark & $+K$ steps \\
\midrule
TextMAS       & \cmark & \cmark & \xmark & \cmark & Decoding \\
\bottomrule
\end{tabular}
\end{table}

Besides its communication interface, a latent multi-agent system differs from a single agent in its harness: an agent scaffold with role-specific prompts, upstream context retained for later agents, and additional computation.
To separate the interface from these factors, we compare five conditions in Figure~\ref{figure:latent_interface_overview} and Table~\ref{table:effect_condition_comparison}.
\emph{Single} and \emph{Judger-Only} answer the question without the multi-agent pipeline.
Judger-Only differs from Single only in using the Judger's role-specific prompt $x^{A_N}$.
It corresponds to LatentMAS at $K=0$, whose implementation discards the upstream cache before final generation.
\emph{Scaffold-Only}, \emph{LatentMAS}, and \emph{TextMAS} employ the multi-agent harness.
Scaffold-Only omits the communication interface among agents.
LatentMAS and TextMAS utilize a latent and text interface, respectively.

Ordered as Single, Judger-Only, Scaffold-Only, and LatentMAS, these conditions form a ladder that adds harness factors cumulatively: role-specific prompt, then the multi-agent scaffold, and finally the latent interface.
TextMAS serves as a reference that communicates through decoded text instead of latent inputs.
Comparing Scaffold-Only with LatentMAS controls for the upstream prompt scaffold and measures the effect of adding the latent interface as a whole.
We treat these steps as controlled comparisons rather than an additive decomposition of system performance.
Moreover, this comparison does not by itself separate the effects of latent content from those of additional input positions or latent-step computation, which the internal level addresses with Zero replacement and latent-step ablations.

%-------------------------------------------------------------------------------
\subsection{Internal Level: Interface Components}
\label{section:internal}
%-------------------------------------------------------------------------------

At the internal level, we examine what happens inside the latent interface.
Recall that at latent step $t$, a hidden state $h_t$ is first realigned to a latent input $r_t = \mathrm{Realign}(h_t)$, which the receiver then processes as $(h_{t+1}, C_{t+1}) = f_\theta(r_t; C_t)$ (see Section~\ref{section:latent_interface}).
Repeating this process allows information and representation-level perturbations to propagate across latent steps.
This formulation highlights an important distinction: \emph{latent communication is not equivalent to effective information transfer.}
A hidden state may encode task-relevant information without providing a useful input to the receiving model.
\method therefore analyzes the latent interface through three interacting components: whether the latent input is compatible with the receiver's input interface (\emph{realignment compatibility}), whether task-relevant information can be used by the receiver (\emph{information accessibility}), and how the effects of latent inputs evolve through repeated computation (\emph{propagation dynamics}).

\mypara{Realignment Compatibility $\mathcal R$}
Realignment compatibility concerns whether a hidden state remains useful when it is transformed into a latent input.
For example, in Equation~\ref{equation:realignment_function}, the realigned $r_t$ might not be in the distribution of input token embeddings.
Note that poor realignment compatibility does not directly imply information loss in hidden states; instead, it is reflected in how the receiver processes the realigned input and, in turn, in downstream performance.

\mypara{Information Accessibility $\mathcal I$}
Previous works assume that information contained in latent representations could enhance reasoning~\cite{HSSLHWT24,ZYQLTLSTCHZWY26,DWBCZCZCY26}.
We thus probe whether task-relevant information is decodable from a latent representation, as a proxy for information accessibility.
Note that information identified by an independently trained probe may still be inaccessible to the receiver through its existing computation.
Preserving semantic information therefore does not necessarily make a latent representation usable.
Limited information accessibility may explain the absence of performance gains, whereas incompatible realignment may additionally introduce harmful perturbations.

\mypara{Propagation Dynamics $\mathcal P$}
Propagation dynamics characterize how the effects of latent inputs evolve over successive computation steps.
As shown in Equation~\ref{equation:latent_step}, because each latent step operates on context produced by previous steps, its behavior depends on the preceding latent trajectory.
Repeated computation may suppress, preserve, or amplify earlier distortions.
Therefore, the effect of a latent interface cannot be inferred solely from the properties of an individual latent input.

\begin{table}[t]
\centering
\caption{Interventions at the internal level of \method, grouped by the component each primarily targets.}
\label{table:internal_interventions}
\setlength{\tabcolsep}{3.3pt}
\begin{tabular}{lp{0.8\columnwidth}}
\toprule
& \textbf{Interventions} \\
\midrule
$\mathcal R$ & Tied vs.\ untied models; LatentMAS w/o $W_a$; Gaussian and dimension-shuffle perturbations before vs.\ after realignment (Section~\ref{section:realignment}) \\
$\mathcal I$ & Domain and subdomain probes on TextMAS hidden states $h^{\text{text}}_j$, LatentMAS hidden states $h_j$, and latent inputs $r_j$ (Section~\ref{section:information}) \\
$\mathcal P$ & Latent steps $K\in\{1,10,20,40,80\}$ with and without $W_a$; Zero replacement (Section~\ref{section:propagation}) \\
\bottomrule
\end{tabular}
\end{table}

Table~\ref{table:internal_interventions} summarizes the interventions targeting each component, which we apply in Section~\ref{section:rq2}.

%-------------------------------------------------------------------------------
\section{RQ1: Decoupling From Agent Harness}
\label{section:rq1}
%-------------------------------------------------------------------------------

%-------------------------------------------------------------------------------
\subsection{Experimental Settings}
%-------------------------------------------------------------------------------

To answer RQ1, we apply the external level of \method (Section~\ref{section:external}) to LatentMAS~\cite{ZYQLTLSTCHZWY26}.
We evaluate Qwen3-4B and Qwen3-8B~\cite{T25} across seven benchmarks: GSM8K~\cite{CKBCJKPTHNHS21}, GPQA~\cite{RHSPPDMB23}, MedQA~\cite{YCGCLHHWW25}, ARC-Easy and ARC-Challenge~\cite{CCEKSST18}, MBPP+~\cite{LXWZ23}, and HumanEval+~\cite{LXWZ23}.
The main comparison uses the sequential Planner-Critic-Refiner-Judger pipeline, with the Judger as the final agent and $K=20$ latent steps per non-final agent, following Zou et al.~\cite{ZYQLTLSTCHZWY26}.
We report task accuracy and its unweighted mean across the seven benchmarks.
Full model, dataset, generation, and computational details are provided in Appendix~\ref{appendix:experimental_details}.

%-------------------------------------------------------------------------------
\subsection{Main Results}
%-------------------------------------------------------------------------------

\begin{table*}[t]
\centering
\caption{Task performance (\%) of the conditions in Table~\ref{table:effect_condition_comparison}.
Avg.\ is the unweighted mean over the seven benchmarks.
HE+ denotes HumanEval+.
$\Delta$ is the descriptive difference.
We conduct McNemar's test for LatentMAS vs.\ Scaffold-Only and mark ``\significant'' when significant (Holm-adjusted \(p<0.05\)).
}
\label{table:main_effects}
\setlength{\tabcolsep}{3.3pt}
\begin{tabular}{lrrrrrrrr}
\toprule
Method & GSM8K & GPQA & MedQA & ARC-E & ARC-C & MBPP+ & HE+ & Avg. \\
\midrule
\multicolumn{9}{l}{\textbf{Qwen3-4B}} \\
Single & 87.79 & 42.42 & \textbf{68.00} & 96.76 & 91.13 & 74.07 & 79.27 & 77.06 \\
Judger-Only ($K=0$) & 88.93 & 41.41 & \textbf{68.00} & 96.97 & 91.64 & 71.16 & 77.44 & 76.51 \\
Scaffold-Only & \textbf{90.67} & \textbf{50.51} & \underline{67.67} & \underline{97.18} & \underline{93.34} & \underline{74.87} & \underline{79.88} & \underline{79.16} \\
TextMAS & \textbf{90.67} & \underline{48.99} & 67.00 & \textbf{97.60} & \textbf{94.03} & \textbf{75.40} & \textbf{90.24} & \textbf{80.56} \\
LatentMAS ($K=20$) & \underline{90.07} & 42.42 & 66.67 & 95.37 & 90.53 & 59.52 & 65.85 & 72.92 \\
$\Delta$(LatentMAS $-$ Scaffold) & -0.61 & -8.08 & -1.00 & -1.81\significant & -2.82\significant & -15.34\significant & -14.02\significant & -6.24 \\
\midrule
\multicolumn{9}{l}{\textbf{Qwen3-8B}} \\
Single & 89.08 & 44.95 & \underline{75.33} & 97.52 & 91.81 & \underline{75.40} & 79.88 & 79.14 \\
Judger-Only ($K=0$) & 89.31 & 46.46 & \underline{75.33} & 97.31 & 91.98 & 72.49 & 80.49 & 79.05 \\
Scaffold-Only & 92.04 & \textbf{56.06} & 74.67 & \underline{98.32} & \underline{94.37} & 74.34 & \underline{81.71} & \underline{81.64} \\
TextMAS & \textbf{93.10} & \underline{52.53} & \textbf{75.67} & \textbf{98.44} & \textbf{95.14} & \textbf{78.84} & \textbf{87.20} & \textbf{82.99} \\
LatentMAS ($K=20$) & \underline{92.27} & 42.93 & 66.00 & 97.90 & 93.94 & 26.46 & 29.88 & 64.20 \\
$\Delta$(LatentMAS $-$ Scaffold) & +0.23 & -13.13\significant & -8.67\significant & -0.42 & -0.43 & -47.88\significant & -51.83\significant & -17.45 \\
\bottomrule
\end{tabular}
\end{table*}

Table~\ref{table:main_effects} shows LatentMAS results with $K=20$ and the corresponding baselines.
Based on our reproduction, non-latent controls such as Scaffold-Only generally outperform LatentMAS across seven benchmark datasets.
For example, the accuracy on Qwen3-4B decreases from $79.16\%$ on Scaffold-Only to $72.92\%$ across seven datasets.
We further conduct McNemar's test for LatentMAS vs.\ Scaffold-Only on each benchmark.
The reduction is significant on four of seven tasks for each model, especially on the code generation tasks MBPP+ and HumanEval+.
We analyze the failure modes of code datasets in Appendix~\ref{appendix:code_failure} and find that LatentMAS shows substantially more invalid or unparsable code outputs.

Further, Judger-Only stays close to Single ($76.51\%$ versus $77.06\%$ for Qwen3-4B and $79.05\%$ versus $79.14\%$ for Qwen3-8B), suggesting that the final agent's role-specific prompt alone contributes little.
Adding the agent scaffold and its upstream context raises the mean to $79.16\%$ and $81.64\%$ for Scaffold-Only, so the harness already improves over Single before any latent input is introduced.
By contrast, adding the latent interface is the only step that substantially lowers mean accuracy.
Notice that Qwen3-4B LatentMAS exceeds Single on GSM8K ($90.07\%$ versus $87.79\%$), yet remains below Scaffold-Only ($90.67\%$).
The same pattern holds for Qwen3-14B~\cite{T25} (Appendix Table~\ref{table:effect_general}).
We thus argue that an improvement of LatentMAS over a single-agent baseline does not, by itself, demonstrate a benefit from the latent interface.

%-------------------------------------------------------------------------------
\subsection{Answer to RQ1}
%-------------------------------------------------------------------------------

When decoupled from the agent harness, the latent interface does not contribute a consistent gain to overall multi-agent performance in the evaluated settings.
Along the ladder, the improvement over Single comes from the agent scaffold and its upstream context (Scaffold-Only) for both models, whereas adding latent communication on top of the same harness reduces mean accuracy by $6.24\%$ and $17.45\%$ for Qwen3-4B and 8B, respectively.
The losses are significant on four of seven tasks for each model, with the largest on code generation, and the same pattern holds for Qwen3-14B (Appendix Table~\ref{table:effect_general}).
RQ2 examines the components within the interface that underlie this outcome.

%-------------------------------------------------------------------------------
\section{RQ2: Decoupling Components within the Latent Interface}
\label{section:rq2}
%-------------------------------------------------------------------------------

To answer RQ2, we apply the internal level of \method (Section~\ref{section:internal}) to Qwen3-4B and Qwen3-8B, examining each component with the interventions in Table~\ref{table:internal_interventions}.

%-------------------------------------------------------------------------------
\subsection{Realignment Compatibility $\mathcal R$}
\label{section:realignment}
%-------------------------------------------------------------------------------

We first examine whether realignment makes hidden states more useful as latent inputs.
We analyze the alignment properties of tied and untied models, then examine how realignment affects task performance through ablations on the hidden states and latent inputs.

%-------------------------------------------------------------------------------
\subsubsection{Tied and Untied Interfaces}
\label{section:tied_untied}
%-------------------------------------------------------------------------------

\begin{table}[t!]
\centering
\caption{Input--output alignment analysis across tied and untied models.
Mean Cosine is the average cosine similarity between corresponding rows of $W_{\mathrm{in}}$ and $W_{\mathrm{out}}$.
Relative Error is $\|W_{\mathrm{out}}W_a-W_{\mathrm{in}}\|_F/\|W_{\mathrm{in}}\|_F$.}
\label{table:matrix_alignment}
\setlength{\tabcolsep}{3.3pt}
\begin{tabular}{lrr}
\toprule
Model & Mean Cosine & Relative Error \\
\midrule
\multicolumn{3}{l}{\textbf{Tied}} \\
Qwen3-4B & 1.000 & $1.4\times10^{-7}$ \\
Llama-3.2-3B-Instruct & 1.000 & $3.2\times10^{-7}$ \\
\midrule
\multicolumn{3}{l}{\textbf{Untied}} \\
Qwen3-8B & 0.004 & 0.919 \\
Qwen3-14B & 0.002 & 0.919 \\
Llama3.1-8B-Instruct & 0.006 & 0.938 \\
\bottomrule
\end{tabular}
\end{table}

Models differ in whether their input embedding matrix and output head share parameters.
In \emph{tied} models, $W_{\mathrm{in}} = W_{\mathrm{out}}$, so that their input and output embeddings are already aligned.
Meanwhile, \emph{untied} models parameterize the two matrices separately, which have near-zero mean cosine similarity.
As a result, realignment may exhibit different compatibility between tied and untied models.
We evaluate this through the relative reconstruction error of $W_a$ in LatentMAS: $\frac{
\left\|W_{\mathrm{out}}W_a-W_{\mathrm{in}}\right\|_F
}{
\left\|W_{\mathrm{in}}\right\|_F
}$.

Table~\ref{table:matrix_alignment} shows the results on Qwen3 and Llama3 models.
Results show a consistent contrast across the evaluated models: reconstruction errors are negligible for tied models but remain substantial for untied models.
This demonstrates that the LatentMAS realignment does not closely recover the input embeddings in the evaluated untied settings.
Note that our primary models provide these two settings: Qwen3-4B is tied, while Qwen3-8B is untied.
The comparison between Qwen3-4B and 8B therefore provides a natural contrast in realignment compatibility: Qwen3-4B is well-realigned, whereas Qwen3-8B is not.
However, poor realignment compatibility does not establish information loss or reasoning impairment.
Thus, we analyze the effect of realignment compatibility via ablations.

%-------------------------------------------------------------------------------
\subsubsection{Experimental Settings}
%-------------------------------------------------------------------------------

To examine the effect of realignment compatibility, we compare Qwen3-4B and 8B models with interventions on the saved states before and after realignment.
Before/After realignment indicates where the ablation happens: Before Realignment applies realignment on the intervened representations, whereas After Realignment directly takes the intervention as input.
Specifically, we consider the following conditions and compare them with Scaffold-Only and LatentMAS with $K=20$:
(i) LatentMAS w/o $W_a$ is an ablation compared with LatentMAS, which removes the transformation matrix $W_a$ in realignment and thus uses the normalized hidden states as inputs;
(ii) Zero uses zero vectors to replace the latent inputs;
(iii) Norm-Matched Gaussian replaces the representation with random noise of the same norm;
and (iv) Dimension Shuffle permutes the dimensions of the representations, either hidden states or latent inputs.

As these ablations reuse the saved states, we further validate that the original LatentMAS is compatible with a replayed version without any ablation in Appendix~\ref{appendix:replay_validation}.
This confirms that the replayed states are valid for evaluating ablations.

%-------------------------------------------------------------------------------
\subsubsection{Effects of Realignment Compatibility}
%-------------------------------------------------------------------------------

\begin{table}[t!]
\centering
\caption{Effects of realignment on mean task performance (\%) across seven benchmarks.
Before/After Realignment indicates when the ablation applies.
}
\label{table:realignment_performance}
\setlength{\tabcolsep}{3.3pt}
\begin{tabular}{lrr}
\toprule
Condition & Qwen3-4B & Qwen3-8B \\
\midrule
Scaffold-Only & \textbf{79.16} & 81.64 \\
LatentMAS & 72.92 & 64.20 \\
LatentMAS w/o $W_a$ & 73.69 & 81.47 \\
Zero & 78.93 & \textbf{82.59} \\
\midrule
\multicolumn{3}{l}{\textbf{Norm-Matched Gaussian}} \\
Before Realignment & 42.44 & 76.54 \\
After Realignment & 43.18 & 81.43 \\
\midrule
\multicolumn{3}{l}{\textbf{Dimension Shuffle}} \\
Before Realignment & 46.49 & 77.50 \\
After Realignment & 45.39 & 81.98 \\
\bottomrule
\end{tabular}
\end{table}

Table~\ref{table:realignment_performance} reports the mean task accuracy; Appendix~\ref{appendix:full_task_results} provides full results.

\mypara{Tied and Untied Models}
Results show that LatentMAS on Qwen3-8B drops more in performance than on Qwen3-4B.
For example, LatentMAS on Qwen3-8B decreases accuracy by 47.88\% on MBPP+, while that on Qwen3-4B decreases by 15.34\%.
Even the tied Qwen3-4B remains below Scaffold-Only, indicating that aligned input and output embeddings do not suffice to make latent inputs beneficial.
We further replace $W_a$ with the identity matrix $I$ to ablate the influence of realignment.
Table~\ref{table:realignment_performance} shows that for the tied Qwen3-4B model, removing $W_a$ changes mean accuracy only slightly, consistent with its near-identity realignment matrix.
For the untied Qwen3-8B model, removing $W_a$ raises accuracy from $64.20\%$ to $81.47\%$, close to Scaffold-Only at $81.64\%$.
This recovery supports the idea that poor realignment compatibility impairs performance in this model.
However, because removing $W_a$ also changes subsequent hidden states and context, this comparison measures its effect throughout the pipeline rather than an isolated transformation of a fixed state.

\mypara{Before and After Realignment}
We further test how realignment affects performance when the inputs are random perturbations.
As shown in Table~\ref{table:realignment_performance}, both perturbations substantially degrade Qwen3-4B relative to the original latent sequence, and applying them before or after realignment yields similar mean accuracies, consistent with $W_a$ being close to identity.
For Qwen3-8B, in contrast, ablations after realignment yield higher accuracy than those applied before it.
Gaussian replacement improves from $76.54\%$ before realignment to $81.43\%$ after it, while dimension shuffling improves from $77.50\%$ to $81.98\%$.
Both after-realignment conditions largely close the gap to Scaffold-Only despite disrupting the original latent inputs.
The Gaussian comparison further suggests that realignment can affect performance even when the injected vectors do not preserve the original task-specific state content.

These contrasting responses show that the effect of randomization is model-dependent: it substantially degrades performance in Qwen3-4B but alleviates the original deficit in Qwen3-8B when applied after realignment.
Since all these interventions retain the injected positions, the differences arise from the injected vectors rather than from the positions themselves, which the Zero control in Section~\ref{section:propagation} examines further.
The before- and after-realignment conditions use separately generated random draws, so their differences should be interpreted as comparisons between intervention procedures rather than paired transformations of identical perturbations.

%-------------------------------------------------------------------------------
\subsection{Information Accessibility $\mathcal I$}
\label{section:information}
%-------------------------------------------------------------------------------

To examine information accessibility in the latent interface, we probe how much information about a question's domain can be decoded from hidden states and realigned latent inputs, using the hidden states of TextMAS as a text reference.
Contrasting coarse domain labels with finer subdomain labels further indicates how specific the decodable information is.
Note that these probes measure this aspect of the representations rather than their overall information content or usefulness for reasoning.

%-------------------------------------------------------------------------------
\subsubsection{Experimental Settings}
%-------------------------------------------------------------------------------

We analyze Qwen3-4B and 8B using states saved from LatentMAS ($K=20$) and TextMAS runs.
For each question and agent, we compare LatentMAS hidden states $h_j$, their realigned latent inputs $r_j$, and TextMAS hidden states $h^{\text{text}}_j$ predicting the $j+1$-th output token.
We evaluate the first state ($h^{\text{text}}_0$, $h_0$, or $r_0$) and the concatenation of 20 states in order ($h^{\text{text}}_{0-19}$, $h_{0-19}$, or $r_{0-19}$).
Although $h^{\text{text}}_0$ and $h_0$ are the last prompt-position hidden states, they are different because the agent's prompt and context are different.
For each model, agent, and representation, we train an L2-regularized logistic-regression probe.
We use nested cross-validation with five outer folds for evaluation and three inner folds for selecting the regularization parameter by accuracy.
We use Zero as the baseline, which supplies all-zero feature vectors to the same classifier under the same evaluation protocol.
All conditions share the same data splits.

We use GPQA's official domain (Physics, Chemistry, and Biology) and subdomain labels to probe coarse- and fine-grained domain information~\cite{RHSPPDMB23}, with details provided in Appendix~\ref{appendix:dataset_details}.

We report Macro-F1 of pooled outer-fold predictions, giving equal weight to each class.

%-------------------------------------------------------------------------------
\subsubsection{Results of Information Accessibility}
%-------------------------------------------------------------------------------

\begin{table}[t!]
\centering
\caption{
Domain and subdomain classification Macro-F1 ($\%$) on GPQA.
}
\label{table:information_classification}
\setlength{\tabcolsep}{3.3pt}
\begin{tabular}{lrrrrrr}
\toprule
& \multicolumn{3}{c}{Domain} & \multicolumn{3}{c}{Subdomain} \\
\cmidrule(lr){2-4}
\cmidrule(lr){5-7}
 & Planner & Critic & Refiner & Planner & Critic & Refiner \\
\midrule
Zero & 21.67 & 21.67 & 21.67 & 6.08 & 6.08 & 6.08 \\
\midrule
\multicolumn{7}{l}{\textbf{Qwen3-4B}} \\
$h^{\text{text}}_0$ & \underline{90.23} & \textbf{79.75} & \textbf{85.67} & \underline{46.31} & 30.63 & 23.20 \\
$h_0$ & 85.67 & 76.08 & 68.05 & 43.78 & \textbf{37.92} & 26.28 \\
$r_0$ & 85.67 & 76.08 & 68.06 & 43.71 & \underline{37.65} & \underline{26.51} \\
$h^{\text{text}}_{0-19}$ & \textbf{94.52} & \underline{78.57} & \underline{81.11} & \textbf{50.37} & 22.63 & \textbf{30.06} \\
$h_{0-19}$ & 79.35 & 63.60 & 58.49 & 25.83 & 22.93 & 21.06 \\
$r_{0-19}$ & 81.40 & 65.80 & 61.34 & 27.99 & 20.40 & 20.27 \\
\midrule
\multicolumn{7}{l}{\textbf{Qwen3-8B}} \\
$h^{\text{text}}_0$ & 88.96 & \textbf{82.34} & \textbf{76.69} & 43.72 & 29.87 & \textbf{32.58} \\
$h_0$ & 88.95 & 74.07 & 58.12 & \underline{47.12} & \textbf{35.28} & 23.54 \\
$r_0$ & \underline{90.71} & 72.11 & 60.96 & 43.86 & 29.36 & 20.87 \\
$h^{\text{text}}_{0-19}$ & \textbf{94.67} & \underline{81.47} & \underline{76.50} & \textbf{62.11} & \underline{33.34} & \underline{31.47} \\
$h_{0-19}$ & 61.76 & 49.94 & 44.83 & 22.58 & 15.65 & 11.99 \\
$r_{0-19}$ & 51.73 & 39.42 & 41.06 & 17.51 & 14.47 & 9.00 \\
\bottomrule
\end{tabular}
\end{table}

Table~\ref{table:information_classification} reports the probe results.
All representations outperform the Zero baseline, so domain information is decodable from each of them; we therefore focus on the gap between text and latent representations.

At the first state, latent representations are close to text.
Averaged over the three agents, $h_0$ is about $9$ points below $h^{\text{text}}_0$ in domain Macro-F1 for both models and on par with it in subdomain Macro-F1 ($35.99$ vs.\ $33.38$ for Qwen3-4B and $35.31$ vs.\ $35.39$ for Qwen3-8B).

The gap widens as states accumulate, because text representations keep their decodability while latent representations lose it.
For example, from the first state to 20 states, Qwen3-8B Planner's subdomain Macro-F1 rises from $43.72$ to $62.11$ for text but falls from $47.12$ to $22.58$ for latent hidden states.
Averaged over agents, the domain gap grows to $17.59$ (Qwen3-4B) and $32.04$ (Qwen3-8B) points, and the subdomain gap, absent at the first state, opens to $11.08$ and $25.57$.
Finer-grained information also fades faster within the latent trajectory.
Measured as Macro-F1 above the Zero baseline, $h_{0-19}$ keeps $83\%$ and $59\%$ of the domain signal of $h_0$ but only $57\%$ and $36\%$ of its subdomain signal.
For Qwen3-8B, $r_{0-19}$ reaches only $9.00$--$17.51$ subdomain Macro-F1, against $6.08$ for Zero.
These results suggest that latent representations retain mainly coarse, domain-level information and little identifiable task-specific information, so the latent interface may carry too little specific information to benefit the receiver.

Realignment explains only part of this decline.
For Qwen3-8B, $r_{0-19}$ falls below $h_{0-19}$ for every agent and label set, consistent with its incompatible realignment (Section~\ref{section:realignment}).
In the tied Qwen3-4B, however, whose realignment is near-identity, $h_{0-19}$ and $r_{0-19}$ decline by similar amounts.
The decline therefore also arises in the hidden states themselves, which we relate to propagation in Section~\ref{section:propagation}.

%-------------------------------------------------------------------------------
\subsection{Propagation Dynamics $\mathcal{P}$}
\label{section:propagation}
%-------------------------------------------------------------------------------

\begin{figure}[t!]
\centering
\includegraphics[width=\columnwidth]{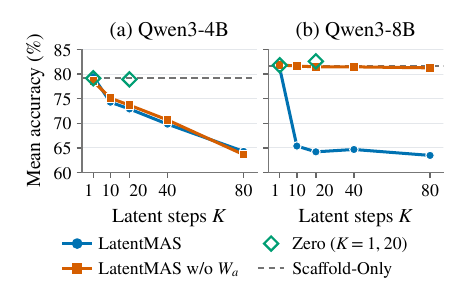}
\caption{
Effects of latent steps $K$, shown as mean accuracies (\%) across seven benchmarks.
}
\label{figure:latent_steps}
\end{figure}

Each latent step updates both the hidden state and the retained context, which jointly influence subsequent computation.
We therefore examine whether increasing the number of latent steps amplifies distortion, comparing LatentMAS with and without $W_a$ across latent steps $K$ (Figure~\ref{figure:latent_steps}; Appendix Table~\ref{table:latent_steps_full}).
For the untied Qwen3-8B, this comparison is dominated by realignment rather than propagation: with $W_a$, mean accuracy falls from $81.22\%$ at $K=1$ to $65.38\%$ at $K=10$ and then plateaus ($63.47\%$ at $K=80$), whereas without $W_a$ it stays between $81.27\%$ and $81.86\%$ for all $K$, consistent with Section~\ref{section:realignment}.
To separate propagation from realignment, we therefore focus on the tied Qwen3-4B, whose realignment is near-identity.

\mypara{Negative Effects Beyond Realignment}
In contrast to this step change, Qwen3-4B degrades gradually.
With $W_a$, mean accuracy decreases from $79.45\%$ at $K=1$ to $72.92\%$ at $K=20$ and $64.29\%$ at $K=80$.
Without $W_a$, it likewise decreases from $78.56\%$ to $73.69\%$ and $63.64\%$, respectively.
The reductions from $K=1$ to $K=80$ are therefore similar: $15.16$ and $14.92\%$.
This behavior is consistent with the near-identity realignment matrix of the tied Qwen3-4B model.
More importantly, removing $W_a$ does not prevent the degradation, showing that a nontrivial linear realignment is not necessary for prolonged latent feedback to be harmful.
Repeatedly processing hidden states as inputs can alter the evolving context in ways that impair subsequent generation, even when the input and output embedding matrices are already aligned.
These results support a negative effect of extended latent propagation and are consistent with harmful deviations being carried forward or amplified.
They do not, however, directly measure per-step distortion or establish that distortion increases at every step.

\mypara{Injected Content Matters}
The Zero controls help distinguish these effects from merely adding input positions.
At $K=20$, Zero achieves $78.93\%$ on Qwen3-4B, close to Scaffold-Only at $79.16\%$.
This contrasts with the degradation under both latent conditions.
Thus, within the tested controls, the observed losses are not explained by the number of injected positions alone; they depend on the injected vectors and their interaction with subsequent computation.

\mypara{Information Loss Along the Trajectory}
This degradation parallels the information loss in Section~\ref{section:information}.
For Qwen3-4B, decodability declines from $r_0$ to $r_{0-19}$, with agent-averaged domain Macro-F1 falling from $76.60$ to $69.51$ and subdomain Macro-F1 from $35.96$ to $22.89$ (Table~\ref{table:information_classification}).
Later latent inputs therefore carry less question-specific information, yet the normalization in realignment, applied with or without $W_a$, keeps them at the norm of an input embedding.
Such inputs approach the norm-preserving perturbations in Section~\ref{section:realignment}, which reduce Qwen3-4B to $42.44$--$46.49\%$, whereas Zero inputs, which inject no signal, leave it close to Scaffold-Only.
Propagation may thus harm Qwen3-4B by gradually turning latent inputs into vectors that carry little information but still have the scale of input embeddings.
This account is consistent with the observed trends, but we do not measure the information carried by each step, so it remains a hypothesis.

%-------------------------------------------------------------------------------
\subsection{Answer to RQ2}
%-------------------------------------------------------------------------------

The internal analyses locate where the RQ1 deficit arises within the interface, with the three components contributing differently.
First, the additional input positions are not harmful by themselves: Zero replacement, which keeps every injected position, stays close to Scaffold-Only in both models (Table~\ref{table:realignment_performance}), so the losses arise from the injected vectors and how they are processed.
On the benefit side, latent representations retain coarse domain information, close to text at the first state.
However, as latent states accumulate, the gap to text widens and finer-grained information fades faster, so the latent interface carries little fine-grained information (Section~\ref{section:information}).
The remaining limitations arise from realignment and propagation.
For the untied Qwen3-8B, incompatible realignment largely accounts for the deficit, which appears as a step change by $K=10$ rather than accumulating over latent steps (Sections~\ref{section:realignment} and~\ref{section:propagation}).
In the tied Qwen3-4B, where realignment is near-identity and propagation can thus be isolated, accuracy degrades gradually with the number of latent steps, from $79.45\%$ at $K=1$ to $64.29\%$ at $K=80$, possibly because later latent inputs carry less information while keeping the scale of input embeddings (Section~\ref{section:propagation}).
Alignment alone therefore does not guarantee useful latent inputs.
Hence, the coarse information preserved in latent representations is not sufficient for effective communication, since the interface carries little fine-grained information and its benefit can be outweighed by realignment and propagation failures.

%-------------------------------------------------------------------------------
\section{Related Work}
%-------------------------------------------------------------------------------

\mypara{Latent Reasoning and Communication}
Latent reasoning replaces intermediate text with recurrent continuous states vitrained realignment~\cite{HSSLHWT24,SYZHDH25}.
Multi-agent systems communicate through several such carriers~\cite{PLYCLYPWY24,DWBCZCZCY26,FMZYDOW26}.
LatentMAS combines realignment and recurrent latent steps without training~\cite{ZYQLTLSTCHZWY26}.
RecursiveMAS and DiffMAS jointly optimize latent collaboration, while LaTER couples a LatentMAS-style transition with explicit verification~\cite{YZPQLDJTZBHZ26,YLJYKW26,LWLLQLLHYZ26}.
These works report accuracy or efficiency gains of complete systems without separating the latent interface from the agent harness.

\mypara{Interpreting Latent Reasoning}
Dilgren et al.~\cite{DW26} first test whether latent tokens are necessary, then recover verifiable textual traces when they are used.
Other work reconstructs and intervenes on latent reasoning states, investigating the trajectory and faithfulness~\cite{CBZMLLMN26,DBNFVZ26,AFZP26}.
These studies concern trained within-model reasoning.
We instead study multi-agent systems and focus on latent contributions rather than trajectory.

\mypara{Analyzing Latent Multi-Agent Interfaces}
Closest to our causal motivation, Zhang et al.~\cite{ZE26} replace current-example messages with multiple ablations to separate message presence, example-specific content, and separate-agent value.
Zhang et al.~\cite{ZBK26} similarly distinguish latent-token content from span-position and token-budget effects on VLMs.
We study a complementary locus: \method decouples the latent interface both from the agent harness and into its components.
Accordingly, we identify interface contributions rather than claiming a quantitative causal decomposition or identical failure modes across latent methods.

%-------------------------------------------------------------------------------
\section{Conclusion}
%-------------------------------------------------------------------------------

In this work, we propose \method, a framework that decouples latent interfaces in multi-agent systems at two levels.
At the external level (RQ1), scaffold-controlled evaluation shows that, relative to the same agent harness, the latent interface does not improve and can substantially degrade overall multi-agent performance.
At the internal level (RQ2), \method decomposes the interface into realignment compatibility, information accessibility, and propagation dynamics.
Latent representations retain some domain information, but incompatible realignment largely accounts for the deficit of the untied Qwen3-8B, and extended latent propagation degrades even the tied Qwen3-4B.
Our framework organizes hypotheses about these components rather than providing a quantitative causal decomposition.
Our findings suggest that, beyond preserving information, successful latent interfaces require compatible realignment, accessible information, and stable propagation.

%-------------------------------------------------------------------------------
\section*{Limitations}
%-------------------------------------------------------------------------------

Our empirical study uses LatentMAS as a primary case study.
Trainable latent-communication methods introduce variables beyond realignment, including the training objective and data, as well as the alignment between the learned latent channel and supervised textual chain-of-thought traces.
However, our findings about realignment compatibility and propagation dynamics can motivate the future design for both trained and training-free methods; extending the proposed diagnostics to both methods is an important direction for future work.
Although the main task comparisons use one random seed, we conduct McNemar's tests to assess statistical significance between condition comparisons.
Furthermore, the information probes measure linear decodability of domain labels rather than direct use of the decoded information by a receiver.
Finally, the tied and untied checkpoints also differ in model size and other architectural properties; our within-model realignment interventions reduce, but do not eliminate this source of confounding.

%-------------------------------------------------------------------------------
\section*{Ethical Considerations}
%-------------------------------------------------------------------------------

This work is intended to improve the understanding and reliability of latent communication in multi-agent systems rather than to support deployment in high-stakes settings.
Latent interfaces may make communication less directly inspectable than textual interaction and can propagate errors across agents without an explicit textual trace.
Consequently, systems using such interfaces should retain appropriate monitoring, external evaluation, and human oversight, particularly in safety-critical applications.

Our study uses publicly available datasets and open-weight language models, and does not involve new data collection, human-subject experiments, or personally identifiable information.
We use the datasets only for their intended research and evaluation purposes and will release our implementation upon publication, subject to the licenses and terms of the corresponding models and datasets, to facilitate reproducibility and independent verification.
Further, we disclose our usage of AI in Appendix~\ref{appendix:ai_usage}.

%-------------------------------------------------------------------------------
\bibliographystyle{plain}
\bibliography{normal_generated_py3}
%-------------------------------------------------------------------------------

%-------------------------------------------------------------------------------
\appendix
%-------------------------------------------------------------------------------

%-------------------------------------------------------------------------------
\section{Disclosure of AI Usage}
\label{appendix:ai_usage}
%-------------------------------------------------------------------------------

Throughout the development phase, AI-based tools integrated into VSCode were used to assist with debugging the implementation of our framework.
During manuscript preparation, we used LLMs for language polishing and grammatical corrections.
Notwithstanding this limited use of AI assistance for specific technical and editorial tasks, the main body of the work was carried out by the human authors.

%-------------------------------------------------------------------------------
\section{Realignment Function}
\label{appendix:realignment}
%-------------------------------------------------------------------------------

A \emph{latent interface} instead realigns a hidden state directly into a continuous input, bypassing discrete token selection and input token embedding.
We denote this as $r_t = \mathrm{Realign}(h_t)$, where $h_t$ is the source hidden state and $r_t$ is the realigned vector supplied as the next input embedding.

In this study, we consider a widely used realignment function~\cite{ZYQLTLSTCHZWY26,YZPQLDJTZBHZ26,YLJYKW26,LWLLQLLHYZ26}:
\begin{equation}
r_t = h_t W_a, \text{where}\  W_a \approx  W^{\dagger}_{\mathrm{out}}W_{\mathrm{in}}.
\end{equation}
$W^{\dagger}_{\mathrm{out}}$ is a pseudo-inverse of $W_{\mathrm{out}}$.
This is a conceptual definition.

Formally, $W_a$ is obtained by regularized least squares:
\begin{equation}
\label{equation:realignment}
W_a =
\left(
W_{\mathrm{out}}^\top W_{\mathrm{out}} + \lambda I_d
\right)^{-1}
W_{\mathrm{out}}^\top W_{\mathrm{in}},
\end{equation}
where $\lambda>0$ is the regularization coefficient.
The realigned vector is rescaled using the mean input embedding norm:
\begin{equation}
\label{equation:realigned_hidden}
\begin{aligned}
r_t &=
\alpha
\frac{h_t W_a}
{\max\!\left(\|h_t W_a\|_2,\epsilon\right)}, \\
\alpha &=
\frac{1}{V}
\sum_{v=1}^{V}\|W_{\mathrm{in}}[v]\|_2,
\end{aligned}
\end{equation}
where $\epsilon>0$ ensures numerical stability.

%-------------------------------------------------------------------------------
\section{Models, Datasets, and Experimental Details}
\label{appendix:experimental_details}
%-------------------------------------------------------------------------------

%-------------------------------------------------------------------------------
\subsection{Models}
%-------------------------------------------------------------------------------

Table~\ref{table:model_checkpoints} lists the checkpoints used in the main experiments, generalization study, and embedding-matrix analysis.
Qwen3-4B and Qwen3-8B are the primary models, while Qwen3-14B is used for the generalization experiments.

\begin{table*}[t]
\centering
\caption{Model checkpoints used in the experiments.}
\label{table:model_checkpoints}
\setlength{\tabcolsep}{3.3pt}
\begin{tabular}{llll}
\toprule
Model & HuggingFace & \#Downloads in Sep. 2026 & Experimental Use \\
\midrule
Qwen3-4B & \texttt{Qwen/Qwen3-4B} & 12,345,694 & Main; matrix analysis \\
Qwen3-8B & \texttt{Qwen/Qwen3-8B} & 9,647,139 & Main; matrix analysis \\
Qwen3-14B & \texttt{Qwen/Qwen3-14B} & 6,164,691 & Generalization; matrix analysis \\
Llama-3.1-8B-Instruct & \texttt{meta-llama/Llama-3.1-8B-Instruct} & 5,858,450 & Matrix analysis \\
Llama-3.2-3B-Instruct & \texttt{meta-llama/Llama-3.2-3B-Instruct} & 1,468,032 & Matrix analysis \\
\bottomrule
\end{tabular}
\end{table*}

%-------------------------------------------------------------------------------
\subsection{Datasets}
\label{appendix:dataset_details}
%-------------------------------------------------------------------------------

The seven benchmarks cover three fields:
(i) math and science reasoning: GSM8K~\cite{CKBCJKPTHNHS21}, GPQA~\cite{RHSPPDMB23}, and MedQA~\cite{YCGCLHHWW25}; (ii) commonsense reasoning: ARC-Easy and ARC-Challenge~\cite{CCEKSST18}; (iii) code generation: MBPP+~\cite{LXWZ23} and HumanEval+~\cite{LXWZ23}.
We evaluate the complete split listed in Table~\ref{table:dataset_details} for every condition unless otherwise stated.
The MedQA evaluation uses the 300-example test set distributed with the AgentFlow evaluation data.
The maximum generation budget is shared across compared conditions within each task.

\begin{table*}[t]
\centering
\caption{Dataset splits, evaluation sizes, and maximum generation budgets used in the seven-benchmark evaluation.}
\label{table:dataset_details}
\setlength{\tabcolsep}{3.3pt}
\begin{tabular}{lllrr}
\toprule
Dataset & Dataset Identifier & Split & \#Examples & Max New Tokens \\
\midrule
GSM8K & \texttt{gsm8k/main} & Test & 1,319 & 4,096 \\
GPQA Diamond & \texttt{fingertap/GPQA-Diamond} & Test & 198 & 10,000 \\
MedQA & AgentFlow evaluation copy & Test & 300 & 8,192 \\
ARC-Easy & \texttt{allenai/ai2\_arc/ARC-Easy} & Test & 2,376 & 4,096 \\
ARC-Challenge & \texttt{allenai/ai2\_arc/ARC-Challenge} & Test & 1,172 & 4,096 \\
MBPP+ & \texttt{evalplus/mbppplus} & Test & 378 & 8,192 \\
HumanEval+ & \texttt{evalplus/humanevalplus} & Test & 164 & 8,192 \\
\bottomrule
\end{tabular}
\end{table*}

For domain probing, we use the official GPQA domain and subdomain labels aligned to the question order of our evaluation runs.
We exclude subdomains with fewer than five examples so that every retained class can occur in each outer test fold and each inner validation fold.
The excluded classes are Condensed Matter Physics (one example), Inorganic Chemistry (one example), Optics and Acoustics (one example), and Genetics (four examples).
This leaves 191 questions with three domains and nine subdomains.
Both classification tasks use exactly these questions, and Table~\ref{table:gpqa_domain_details} reports the retained label counts.

\begin{table*}[t!]
\centering
\caption{
Official domain and subdomain label counts in the matched 191-question GPQA subset.
}
\label{table:gpqa_domain_details}
\setlength{\tabcolsep}{3.3pt}
\begin{tabular}{lr}
\toprule
Label & Count \\
\midrule
\multicolumn{2}{l}{\textbf{Domain}} \\
Physics & 84 \\
Chemistry & 92 \\
Biology & 15 \\
\midrule
\multicolumn{2}{l}{\textbf{Subdomain}} \\
Astrophysics & 13 \\
Chemistry (General) & 20 \\
Electromagnetism and Photonics & 6 \\
High-Energy Particle Physics & 14 \\
Molecular Biology & 15 \\
Organic Chemistry & 72 \\
Physics (General) & 19 \\
Quantum Mechanics & 25 \\
Relativistic Mechanics & 7 \\
\bottomrule
\end{tabular}
\end{table*}

%-------------------------------------------------------------------------------
\subsection{Generation and Computation}
%-------------------------------------------------------------------------------

The main experiments use the sequential Planner--Critic--Refiner--Judger pipeline, with $K=20$ latent steps per non-judger agent; the propagation study additionally evaluates $K\in\{1,10,20,40,80\}$.
Generation uses temperature $0.6$, nucleus probability $0.95$, and seed 42; top-$k$ is inherited from each checkpoint's generation configuration.
The core LatentMAS and replay-intervention runs use Hugging Face Transformers in bfloat16 with batch size one on a single NVIDIA A100 GPU.
The language-model weights remain frozen throughout all experiments.
The realignment matrix is solved in float32 with ridge coefficient $\lambda=10^{-5}$, and each realigned vector is rescaled to the mean input-embedding norm with denominator clamp $\epsilon=10^{-6}$.
Replay interventions reconstruct the same prompt-prefill and per-latent-step computation as the source rollout; the unchanged replay control in Appendix~\ref{appendix:replay_validation} verifies exact task-level agreement before interventions are interpreted.

%-------------------------------------------------------------------------------
\section{Full Task-Level Results}
\label{appendix:full_task_results}
%-------------------------------------------------------------------------------

Tables~\ref{table:realignment_performance_full} and~\ref{table:latent_steps_full} expand the seven-task averages reported in the main text into results for each benchmark.

\begin{table*}[t!]
\centering
\caption{Full task-level effects of realignment at $K=20$ (\%).
Pre/Post denotes whether the perturbation is applied before or after realignment; Avg. is the unweighted seven-task mean; HE+ denotes HumanEval+.}
\label{table:realignment_performance_full}
\setlength{\tabcolsep}{3.3pt}
\begin{tabular}{lrrrrrrrr}
\toprule
Condition & GSM8K & GPQA & MedQA & ARC-E & ARC-C & MBPP+ & HE+ & Avg. \\
\midrule
\multicolumn{9}{l}{\textbf{Qwen3-4B}} \\
Scaffold-Only & 90.67 & 50.51 & 67.67 & 97.18 & 93.34 & 74.87 & 79.88 & 79.16 \\
LatentMAS & 90.07 & 42.42 & 66.67 & 95.37 & 90.53 & 59.52 & 65.85 & 72.92 \\
LatentMAS w/o $W_a$ & 88.63 & 47.98 & 64.00 & 94.82 & 90.36 & 61.11 & 68.90 & 73.69 \\
Zero & 89.92 & 46.97 & 65.33 & 97.56 & 92.92 & 73.81 & 85.98 & 78.93 \\
Norm-Gaussian (Pre) & 87.72 & 27.78 & 30.00 & 25.76 & 26.45 & 48.15 & 51.22 & 42.44 \\
Norm-Gaussian (Post) & 87.72 & 25.25 & 32.67 & 24.58 & 26.88 & 50.26 & 54.88 & 43.18 \\
Dim. Shuffle (Pre) & 87.95 & 30.30 & 32.33 & 32.53 & 32.94 & 52.65 & 56.71 & 46.49 \\
Dim. Shuffle (Post) & 87.72 & 28.28 & 34.67 & 33.33 & 30.72 & 52.38 & 50.61 & 45.39 \\
\midrule
\multicolumn{9}{l}{\textbf{Qwen3-8B}} \\
Scaffold-Only & 92.04 & 56.06 & 74.67 & 98.32 & 94.37 & 74.34 & 81.71 & 81.64 \\
LatentMAS & 92.27 & 42.93 & 66.00 & 97.90 & 93.94 & 26.46 & 29.88 & 64.20 \\
LatentMAS w/o $W_a$ & 92.04 & 52.53 & 75.33 & 98.27 & 95.31 & 75.13 & 81.71 & 81.47 \\
Zero & 92.49 & 54.55 & 77.67 & 98.19 & 94.97 & 76.72 & 83.54 & 82.59 \\
Norm-Gaussian (Pre) & 91.96 & 48.99 & 74.33 & 98.44 & 94.54 & 61.64 & 65.85 & 76.54 \\
Norm-Gaussian (Post) & 91.74 & 55.05 & 74.33 & 98.32 & 94.80 & 74.07 & 81.71 & 81.43 \\
Dim. Shuffle (Pre) & 90.90 & 51.01 & 73.67 & 98.57 & 94.88 & 64.55 & 68.90 & 77.50 \\
Dim. Shuffle (Post) & 92.19 & 53.54 & 74.67 & 98.36 & 94.80 & 76.19 & 84.15 & 81.98 \\
\bottomrule
\end{tabular}
\end{table*}

\begin{table*}[t!]
\centering
\caption{Full task-level effects of latent steps $K$ (\%).
$K$ is the number of latent steps per non-judger agent; Avg. is the unweighted seven-task mean; HE+ denotes HumanEval+.}
\label{table:latent_steps_full}
\setlength{\tabcolsep}{3.3pt}
\begin{tabular}{lrrrrrrrr}
\toprule
Condition & GSM8K & GPQA & MedQA & ARC-E & ARC-C & MBPP+ & HE+ & Avg. \\
\midrule
\multicolumn{9}{l}{\textbf{Qwen3-4B}} \\
Scaffold-Only & 90.67 & 50.51 & 67.67 & 97.18 & 93.34 & 74.87 & 79.88 & 79.16 \\
\multicolumn{9}{l}{\emph{LatentMAS}} \\
\quad $K=1$  & 90.07 & 50.00 & 67.67 & 97.26 & 93.17 & 73.81 & 84.15 & 79.45 \\
\quad $K=10$ & 90.07 & 45.45 & 65.67 & 96.76 & 92.15 & 61.64 & 68.29 & 74.29 \\
\quad $K=20$ & 90.07 & 42.42 & 66.67 & 95.37 & 90.53 & 59.52 & 65.85 & 72.92 \\
\quad $K=40$ & 87.95 & 41.92 & 61.00 & 91.29 & 87.63 & 55.03 & 64.02 & 69.83 \\
\quad $K=80$ & 84.31 & 28.28 & 53.67 & 83.21 & 78.58 & 57.94 & 64.02 & 64.29 \\
\multicolumn{9}{l}{\emph{LatentMAS w/o $W_a$}} \\
\quad $K=1$  & 90.75 & 48.48 & 63.67 & 97.31 & 93.43 & 74.60 & 81.71 & 78.56 \\
\quad $K=10$ & 89.61 & 46.46 & 65.33 & 97.01 & 92.83 & 61.90 & 72.56 & 75.10 \\
\quad $K=20$ & 88.63 & 47.98 & 64.00 & 94.82 & 90.36 & 61.11 & 68.90 & 73.69 \\
\quad $K=40$ & 88.93 & 36.36 & 62.00 & 91.50 & 86.35 & 59.52 & 70.12 & 70.68 \\
\quad $K=80$ & 82.56 & 23.74 & 54.67 & 82.45 & 78.24 & 59.79 & 64.02 & 63.64 \\
\multicolumn{9}{l}{\emph{Zero}} \\
\quad $K=1$  & 90.07 & 44.95 & 68.00 & 97.26 & 93.00 & 75.93 & 84.76 & 79.14 \\
\quad $K=20$ & 89.92 & 46.97 & 65.33 & 97.56 & 92.92 & 73.81 & 85.98 & 78.93 \\
\midrule
\multicolumn{9}{l}{\textbf{Qwen3-8B}} \\
Scaffold-Only & 92.04 & 56.06 & 74.67 & 98.32 & 94.37 & 74.34 & 81.71 & 81.64 \\
\multicolumn{9}{l}{\emph{LatentMAS}} \\
\quad $K=1$  & 91.66 & 52.02 & 75.00 & 98.32 & 95.31 & 73.28 & 82.93 & 81.22 \\
\quad $K=10$ & 92.34 & 45.45 & 67.00 & 97.85 & 93.77 & 30.16 & 31.10 & 65.38 \\
\quad $K=20$ & 92.27 & 42.93 & 66.00 & 97.90 & 93.94 & 26.46 & 29.88 & 64.20 \\
\quad $K=40$ & 90.45 & 41.41 & 69.67 & 96.17 & 92.92 & 30.42 & 31.71 & 64.68 \\
\quad $K=80$ & 89.99 & 42.93 & 66.00 & 93.86 & 90.44 & 31.22 & 29.88 & 63.47 \\
\multicolumn{9}{l}{\emph{LatentMAS w/o $W_a$}} \\
\quad $K=1$  & 92.12 & 53.03 & 73.00 & 98.36 & 94.54 & 75.40 & 86.59 & 81.86 \\
\quad $K=10$ & 91.81 & 55.05 & 74.67 & 98.40 & 95.22 & 74.60 & 81.71 & 81.64 \\
\quad $K=20$ & 92.04 & 52.53 & 75.33 & 98.27 & 95.31 & 75.13 & 81.71 & 81.47 \\
\quad $K=40$ & 91.96 & 53.03 & 74.33 & 98.44 & 95.31 & 74.87 & 82.32 & 81.47 \\
\quad $K=80$ & 92.34 & 52.02 & 73.00 & 98.44 & 95.39 & 75.40 & 82.32 & 81.27 \\
\multicolumn{9}{l}{\emph{Zero}} \\
\quad $K=1$  & 92.42 & 54.55 & 77.67 & 98.53 & 94.37 & 73.28 & 81.71 & 81.79 \\
\quad $K=20$ & 92.49 & 54.55 & 77.67 & 98.19 & 94.97 & 76.72 & 83.54 & 82.59 \\
\bottomrule
\end{tabular}
\end{table*}

%-------------------------------------------------------------------------------
\section{Code Generation Failures}
\label{appendix:code_failure}
%-------------------------------------------------------------------------------

\begin{table*}[t!]
\centering
\caption{
Code-generation failures, shown as Scaffold-Only $\rightarrow$ LatentMAS.
Invalid Output includes code-extraction failures and syntax errors.
Execution Failure combines runtime exceptions, assertion failures, and timeouts.}
\label{table:code_failure_summary}
\setlength{\tabcolsep}{3.3pt}
\begin{tabular}{ccrr}
\toprule
Dataset & Model & Invalid Output & Execution Failure \\
\midrule
\multirow{2}{*}{\shortstack{MBPP+\\($378$)}}
& Qwen3-4B & 32$\rightarrow$91 & 63$\rightarrow$62 \\
& Qwen3-8B & 35$\rightarrow$207 & 62$\rightarrow$71 \\
\midrule
\multirow{2}{*}{\shortstack{HE+\\($164$)}}
& Qwen3-4B & 17$\rightarrow$43 & 16$\rightarrow$13 \\
& Qwen3-8B & 16$\rightarrow$105 & 14$\rightarrow$10 \\
\bottomrule
\end{tabular}
\end{table*}

In Table~\ref{table:main_effects}, we observe that the largest reductions occur on code generation tasks, i.e., MBPP+ and HumanEval+, while other benchmarks show smaller changes.
Based on the generated answers and execution logs, we classify failures into invalid outputs and execution failures.
Table~\ref{table:code_failure_summary} shows a substantial increase in invalid outputs under LatentMAS for both models, while the difference in execution failure numbers is small.
For example, on Qwen3-8B, the number of invalid MBPP+ outputs rises from 35 under Scaffold-Only to 207 under LatentMAS.
Unlike other benchmarks scoring a final answer extracted from \texttt{\textbackslash boxed\{FINAL ANSWER\}}, code evaluation requires executable code.
This pattern suggests that such a strict requirement may make these tasks particularly sensitive to distortion introduced by the latent interface.

%-------------------------------------------------------------------------------
\section{Generalizability of RQ1 Findings}
\label{appendix:effect_general}
%-------------------------------------------------------------------------------

Table~\ref{table:effect_general} shows Qwen3-14B performance on text baselines, the Scaffold-Only control, and LatentMAS under different latent steps $K$.
Scaffold-Only ($84.35\%$) lies between Single ($82.22\%$) and TextMAS ($85.67\%$), while LatentMAS at $K=20$ falls below it on all seven benchmarks, each significant under McNemar's test.
Unlike Qwen3-4B and 8B, the losses also reach GSM8K and ARC, where the smaller models stay within about $3\%$ of Scaffold-Only.
The model also exhibits a substantial decrease as latent feedback is extended.
Qwen3-14B is untied (Section~\ref{section:tied_untied}), and its embedding reconstruction error is consistent with realignment contributing to this degradation; the step sweep is descriptive and does not isolate realignment from propagation.

\begin{table*}[t!]
\centering
\caption{Performance (\%) of Qwen3-14B under different settings across seven benchmarks.
$\Delta$ is the descriptive difference between LatentMAS ($K=20$) and Scaffold-Only; ``\significant'' marks significance under McNemar's test (Holm-adjusted \(p<0.05\)), as in Table~\ref{table:main_effects}.}
\label{table:effect_general}
\setlength{\tabcolsep}{3.3pt}
\begin{tabular}{lrrrrrrrr}
\toprule
Condition & GSM8K & GPQA & MedQA & ARC-E & ARC-C & MBPP+ & HE+ & Avg. \\
\midrule
\multicolumn{9}{l}{\textbf{Qwen3-14B}} \\
Single & 90.45 & 51.52 & 80.33 & 97.81 & 93.17 & 75.66 & 86.59 & 82.22 \\
Judger-Only ($K=0$) & 90.75 & 52.02 & 82.33 & 97.73 & 93.69 & 73.54 & 83.54 & 81.94 \\
Scaffold-Only & 92.27 & 57.58 & 80.67 & 98.78 & 96.08 & 77.25 & 87.80 & 84.35 \\
TextMAS & 92.95 & 61.62 & 79.00 & 98.65 & 95.65 & 80.95 & 90.85 & 85.67 \\
LatentMAS ($K=10$) & 89.23 & 46.46 & 68.33 & 97.05 & 94.03 & 55.03 & 54.27 & 72.06 \\
LatentMAS ($K=20$) & 81.88 & 27.78 & 43.33 & 92.55 & 86.86 & 36.77 & 29.27 & 56.92 \\
LatentMAS ($K=40$) & 60.12 & 6.06 & 6.00 & 48.11 & 50.26 & 19.84 & 7.32 & 28.24 \\
LatentMAS ($K=80$) & 32.68 & 0.51 & 0.67 & 5.56 & 4.10 & 3.97 & 3.05 & 7.22 \\
$\Delta$(LatentMAS ($K=20$) $-$ Scaffold) & -10.39\significant & -29.80\significant & -37.33\significant & -6.23\significant & -9.22\significant & -40.48\significant & -58.54\significant & -27.42 \\
\bottomrule
\end{tabular}
\end{table*}

%-------------------------------------------------------------------------------
\section{Saved-State Replay Validation}
\label{appendix:replay_validation}
%-------------------------------------------------------------------------------

Because our ablations reuse saved states from LatentMAS, we validate that replaying the unmodified states reproduces the source rollout.
Table~\ref{table:replay_validation} reports the replay performance of Qwen3-4B and Qwen3-8B without interventions.
Both models reproduce the original LatentMAS task-level performance exactly, confirming that the replayed states are valid for evaluating the effects of ablations.

\begin{table*}[t!]
\centering
\caption{Replay validation for Qwen3-4B and Qwen3-8B at $K=20$, seed 42.
Entries are task accuracy (\%).
Exact task-level agreement is an empirical validation of this replay implementation.}
\label{table:replay_validation}
\setlength{\tabcolsep}{3.3pt}
\begin{tabular}{lrrrrrrrr}
\toprule
Condition & GSM8K & GPQA & MedQA & ARC-E & ARC-C & MBPP+ & HE+ & Avg. \\
\midrule
\multicolumn{9}{l}{\textbf{Qwen3-4B}} \\
LatentMAS & 90.07 & 42.42 & 66.67 & 95.37 & 90.53 & 59.52 & 65.85 & 72.92 \\
Saved-State Replay & 90.07 & 42.42 & 66.67 & 95.37 & 90.53 & 59.52 & 65.85 & 72.92 \\
\midrule
\multicolumn{9}{l}{\textbf{Qwen3-8B}} \\
LatentMAS & 92.27 & 42.93 & 66.00 & 97.90 & 93.94 & 26.46 & 29.88 & 64.20 \\
Saved-State Replay & 92.27 & 42.93 & 66.00 & 97.90 & 93.94 & 26.46 & 29.88 & 64.20 \\
\bottomrule
\end{tabular}
\end{table*}

%-------------------------------------------------------------------------------
\end{document}